\documentclass[twocolumn]{article}

\usepackage{spconf,epsfig}
\usepackage[utf8]{inputenc}
\usepackage[T1]{fontenc}
\usepackage{subcaption}
\usepackage{dsfont}
\usepackage{amsthm}
\usepackage{graphicx}
\usepackage{hyperref}
\usepackage{stackengine}
\usepackage{bm}
\usepackage{amsmath}
\usepackage{soul}
\DeclareMathOperator*{\argmin}{arg\,min}

\newcommand{\nada}[1]{}

\usepackage{tikz}
\usetikzlibrary{spy}
\newcommand{\spied}[2]{\begin{tikzpicture}[every node/.style={inner sep=0,outer sep=0},spy using  outlines={white,magnification=3,size=1.5cm, connect spies}]
\node {\pgfimage[width=\textwidth]{#2}} ;
\spy on #1 in node at (current bounding box.south west) [anchor=south west];
\end{tikzpicture}}

\title{Assessing the Sharpness of Satellite Images:\\Study of the PlanetScope Constellation}

\name{Jérémy Anger$^{\star}$\qquad Carlo de Franchis$^{\star,\dagger}$\qquad Gabriele Facciolo$^{\star}$\thanks{Work partly financed by Office of Naval research  grant N00014-17-1-2552, DGA Astrid project « filmer la Terre » n$^{\circ}$ANR-17-ASTR-0013-01, MENRT; DGA PhD scholarship jointly supported with FMJH.}}

\address{$^{\star}$CMLA, ENS Cachan, CNRS, Université Paris-Saclay, France\qquad\\
$^{\dagger}$Kayrros, France}

\def\sharpness{\mathcal{S}}
\def\basic{\emph{basic}}
\def\ortho{\emph{ortho}}

\begin{document}

\maketitle{}

\begin{abstract}
New micro-satellite constellations enable unprecedented systematic monitoring applications thanks to their wide coverage and short revisit capabilities. However, the large volumes of images that they produce have uneven qualities, creating the need for automatic quality assessment methods.
In this work, we quantify the sharpness of images from the PlanetScope constellation by estimating the blur kernel from each image.
Once the kernel has been estimated, it is possible to compute an absolute measure of sharpness which allows to discard low quality images and deconvolve blurry images before any further processing.
The method is fully blind and automatic, and since it does not require the knowledge of any satellite specifications it can be ported to other constellations. 
\end{abstract}

\section{Introduction}

The usability of satellite images for interpretation, or object detection and reconstruction purposes highly depends on the image quality, which can be characterized by a large number of measures, e.g. contrast, brightness, noise variance, radiometric resolution, sharpness, etc. Among those measures, image sharpness is one of the most important for characterizing images as it evaluates image blur, which limits the visibility of details. 
Image blur is introduced by both the optical system and potential motion during the acquisition time~\cite{Choi2005}.

Assuming a stationary blur kernel $k$ (or \emph{Point Spread Function} (PSF)) that combines the optical and motion blur we can formulate the image formation model as
\begin{equation}
v = u\ast{}k + n,\label{eq:problem}
\end{equation}
where $v$ is the blurry image, $u$ is the latent sharp image, and $n$ is acquisition noise.
Then, sharpness can be objectively measured by estimating the point spread function $k$ or its amplitude spectrum, the \emph{Modulation Transfer Function} (MTF).

In remote sensing, most sharpness studies~\cite{Blanc2009,pagnutti2010targets} focus on the in-flight characterization of the camera system. These approaches usually rely on the presence of on-ground targets such as edges, lines, or point reflectors.
Some methods estimate the blur kernel assuming a parametric model, for example Gaussian, and try to fit its parameters~\cite{Josiane2002}.
Another group of methods estimate cross sections of the MTF, usually by applying some variant of the slanted edge method over calibration sites~\cite{Blanc2009}. These methods allow to periodically assess the sharpness of the system, however they cannot account for motion blur of a particular acquisition, an artifact that has become more common within modern fleets of micro-satellites. 

Other methods~\cite{kohm2004modulation,Crespi2009,luxen2002characterizing} seek to estimate the MTF by relying on the detection of straight edges naturally present in the scene. This allows to estimate sharpness (mostly on urban scenes) without the need of a calibration target. 
Note however that the MTF is only useful for characterizing the sharpness and, while it allows a simple frequency enhancement, it cannot be used to restore the image as the phase of the kernel is not estimated.

Estimating the blur kernel from a single image is an active field of research, especially for natural images since it is a necessary step of most blind deblurring methods~\cite{Levin2009,Shan2008a,Pan2017}. These methods rely only on the presence of contrasted edges (not necessarily straight lines), which allows to apply them to virtually any scene.  
In addition to estimating the kernel for quality assessment, one may want to restore the sharp image $u$.
Indeed, if two images have different blur kernels, it might be difficult to compare and analyze, both visually and automatically by advanced image processing techniques.

We focus our experiments on PlanetScope~\cite{planet} images. The PlanetScope constellation is made of approximately 130 small satellites (form factor of $10\times 10\times 30$ cm) imaging the entire Earth's landmass every day. The satellites carry 3-band or 4-band frame cameras and fly at 475 km on sun-synchronous orbits whose constant local solar time is between 9:30 and 11:30 am. The Ground Sample Distance (GSD) at nadir is between 3.5 m and 4 m. The images are available as either individual \emph{basic} scenes, \emph{ortho} scenes, or \emph{ortho} tiles. We focus here on the \emph{basic} and \emph{ortho} scenes, respectively for non-orthorectified and orthorectified images.
Exploiting both products allows us to infer the influence of the orthorectification on the image sharpness.

\vspace{0.5em}\noindent\textbf{Contributions.~}
We propose a criteria to assess the sharpness of satellite images through the estimation of their blur kernels.
This criteria allows to sort images by quality, thus giving an absolute threshold to discard low quality images and allowing to increase the quality of blurry images using a deblurring step.
Our method is fully blind and is designed for consumers of PlanetScope images. Indeed it does not require the precise specifications of the satellites and could also be used for images from other sources.
We validate our methodology through a study of the PlanetScope constellation. In particular we show the effect of the orthorectification on the sharpness and we study the per-satellite sharpness.

\section{Sharpness assessment and deblurring}

In this section we detail our methodology by first explaining how to blindly estimate the blur kernel and compute a sharpness score from it,  then describe the deblurring step.

\subsection{Blur kernel estimation}

In this work, we use the kernel estimation method of Pan et al.~\cite{Pan2017} originally developed to deblur text images.
This method is based on the $\ell_0$ gradient prior which restores the main structures of the image, including dominant edges.
Once the edges are restored, the blur kernel can be estimated.
The method iterates between the estimation of the sharp image using the previous kernel and the re-estimation of the blur kernel
\begin{align}
u^{(t+1)} &= \argmin_u \|u\ast{}k^{(t)} - v\|_2^2 + \lambda \|\nabla u\|_0
\\k^{(t+1)} &= \argmin_k \|\nabla u^{(t+1)}\ast{}k - \nabla v\|_2^2 + \gamma \|k\|_2^2.
\end{align}
We use the efficient implementation of Anger et al.~\cite{l0ipol}.

Even if the $\ell_0$ gradient prior kernel estimation method was designed for text and natural images, we argue that it is applicable to satellite images without any adaptation.
Indeed, the assumption behind this prior is that non-blurry images contain contrasted edges, which is valid for satellite images. %
Furthermore, satellite image are more likely to respect the stationary convolution model than natural images since the scene is far away from the camera, which results in less parallax, a mostly translational motion and low optical distortion.

\subsection{Measure of sharpness}
\label{sec:sharpness}

\begin{figure}
	\centering
	\def\s{0.19\linewidth}
	\begin{subfigure}{\s}
		\includegraphics[width=\linewidth]{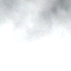}
	\end{subfigure}
	\begin{subfigure}{\s}
		\includegraphics[width=\linewidth]{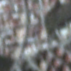}
	\end{subfigure}
	\begin{subfigure}{\s}
		\includegraphics[width=\linewidth]{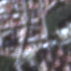}
	\end{subfigure}
	\begin{subfigure}{\s}
		\includegraphics[width=\linewidth]{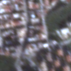}
	\end{subfigure}
	\begin{subfigure}{\s}
		\includegraphics[width=\linewidth]{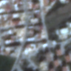}
	\end{subfigure}
	\\[0.13em]
	\begin{subfigure}{\s}
		\includegraphics[width=\linewidth]{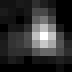}
		\caption{0.019}\label{subfig:clouds}
	\end{subfigure}
	\begin{subfigure}{\s}
		\includegraphics[width=\linewidth]{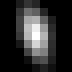}
		\caption{0.022}
	\end{subfigure}
	\begin{subfigure}{\s}
		\includegraphics[width=\linewidth]{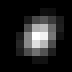}
		\caption{0.026}
	\end{subfigure}
	\begin{subfigure}{\s}
		\includegraphics[width=\linewidth]{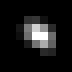}
		\caption{0.032}
	\end{subfigure}
	\begin{subfigure}{\s}
		\includegraphics[width=\linewidth]{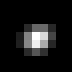}
		\caption{0.033}
	\end{subfigure}
	
    \vspace{-0.5em}
	\caption{Samples of estimated blur kernels. The top rows shows crops of orthorectified images with their associated sharpness score on the bottom row. Clouds mislead the estimation towards a low sharpness score}\label{fig:kernels-and-scores}
\end{figure}

Existing quality assessment metrics for satellite images include measures on the PSF or on the MTF, please refer to Blanc et al.~\cite{Blanc2009} for a comprehensive study. 
We design our sharpness score so that the maximal score of $1$ is achieved for a perfectly sharp image (delta kernel) and it decreases for blurrier images (spread out kernels). Let us note that the kernel is assumed to be normalized so that $\|k\|_1=1$.
The simplest measure satisfying these criteria is the $\ell_2$ norm
\begin{equation}
	\sharpness = \|k\|_2 = \sqrt{{\textstyle\sum}_{\bm x} |k(\bm x)|^2}.
\end{equation}

An advantage of using the blur kernel to assess sharpness is that it is independent of the image content, which is not the case for measures estimated based on properties of the image itself.
Thus the resulting $\sharpness$ score is absolute and can be compared across scenes and/or satellites.
While very simple, this measure is sufficient to characterize the quality of satellite images for many applications. 

Figure~\ref{fig:kernels-and-scores} shows five crops of PlanetScope orthorectified images and their associated kernel and $\sharpness$ score.
We observe that ordering the images by their estimated sharpness indeed correlates well with our perception of the blur introduced by the respective kernel.
Furthermore, we observe that when the scene contains a majority of clouds, the kernel estimation tends to give a very spread out kernel.
This phenomenon occurs because clouds do not have sparse gradients and thus hinder the estimation.
Fortunately, most of the time the predicted sharpness of cloudy scenes is very low, allowing to sort such images as low quality and discard them.

\subsection{Satellite image deblurring}

\begin{figure}
	\centering
	\def\pos{(0,1)}
	\begin{subfigure}{0.46\linewidth}
		\stackinset{r}{}{b}{}{\includegraphics[width=1.2cm]{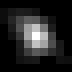}}{\spied{\pos}{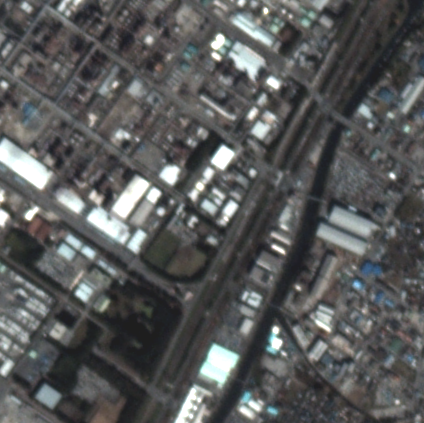}}
	\end{subfigure}
	\begin{subfigure}{0.46\linewidth}
		\stackinset{r}{}{b}{}{\includegraphics[width=1.2cm]{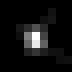}}{\spied{\pos}{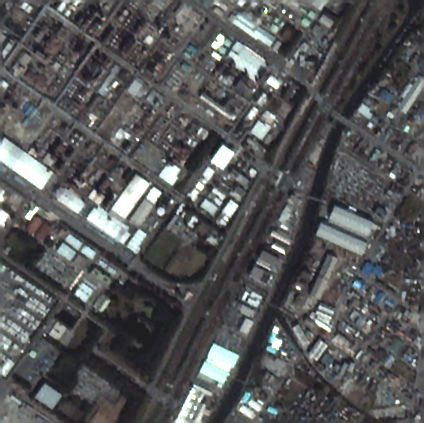}}
	\end{subfigure}
    \vspace{-.4em}
	\caption{Deblurring of an orthorectified image of Tokyo. The input image (left, $\sharpness=0.023$) contains motion blur that is removed after deconvolution (right, $\sharpness=0.035$).}\label{fig:deconvolution-result}
\end{figure}

Having estimated a blur kernel, it is possible to inverse the problem~\eqref{eq:problem} using non-blind deconvolution methods~\cite{Rudin1994,Krishnan2009}.
Since satellite images usually contain low noise in nominal conditions, a simple prior is enough to recover a high quality image.
In this work, we use the total variation prior (TV)~\cite{Rudin1994} already well studied for satellite image deblurring~\cite{Durand2000}, leading to the following optimization problem
\begin{equation}
\argmin_u \|u\ast{}k - v\|_2^2 + \alpha \|\nabla u\|_1.
\label{eq:non-blind-deconvolution-tv}
\end{equation}
We solve problem~\eqref{eq:non-blind-deconvolution-tv} using the fast method from Krishnan et al.~\cite{Krishnan2009}.
Figure~\ref{fig:deconvolution-result} shows a deconvolution result on an orthorectified image. Notice that the input image contains an anisotropic blur, successfully removed by deblurring.

\section{Study of the PlanetScope constellation}

In this section we apply our sharpness measure on PlanetScope images to show that it captures the variability present in the images from this constellation.
As dataset, we collected $7600$ \basic{} and $6790$ \ortho{} PlanetScopes images at 29 different locations for a total of $14390$ images.

\begin{figure}
\centering
\includegraphics[width=.9\linewidth]{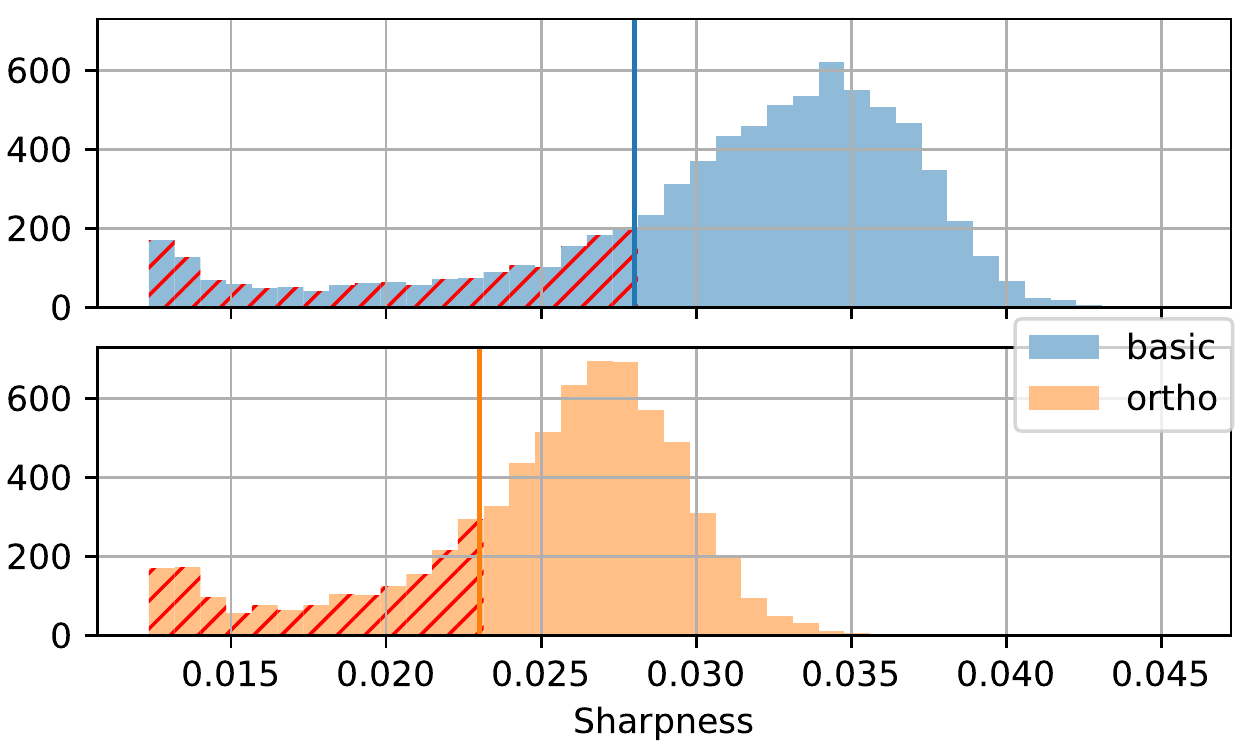}
\vspace{-0.5em}
\caption{Histogram of sharpness for the dataset of $14390$ images. The hatched region represents low quality images or very cloudy scenes.}\label{fig:histogram}
\end{figure}

\vspace{0.5em}\noindent\textbf{Sharpness distribution.~}
Figure~\ref{fig:histogram} shows two histograms representing the distribution of sharpness for \basic{} and \ortho{} images.
We note that the sharpness of \emph{ortho} images is on average lower and less variable than the one of \emph{basic} images.
Indeed, the sharpness after orthorectification decreases significantly, with an average of $\sharpness=0.0251$ versus $\sharpness=0.0309$ for \basic{} images.
This indicates that the orthorectified images are indeed less sharp and our measure does quantify the amount of sharpness lost due to the resampling.

We also observe that both distributions have a second mode near $\sharpness=0.01$. This mode correspond to invalid kernels which can occur on very cloudy scenes for example (as shown on Figure~\ref{subfig:clouds}), or when the signal to noise ratio (SNR) is low due to poor atmospheric conditions.

\vspace{0.5em}\noindent\textbf{Quality thresholds.~}
From these histograms and our observations of the data, we found that for orthorectified images, the threshold $\sharpness>0.030$ indicates high sharpness images whereas $\sharpness<0.023$ corresponds to highly blurred images leaving little hope for a high quality restoration (in red on Figure~\ref{fig:histogram}).
Otherwise, the image can be sharpened using the previously described deblurring algorithm in order to increase its quality before visualization or processing.
For \basic{} images, $\sharpness<0.028$ provides a similar threshold while accounting for the increase of sharpness compared to \ortho{} images.

\begin{figure}
\centering
\def\pos{(0,1)}
\begin{subfigure}{0.46\linewidth}
	\stackinset{r}{}{b}{}{\includegraphics[width=1.2cm]{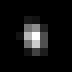}}{\spied{\pos}{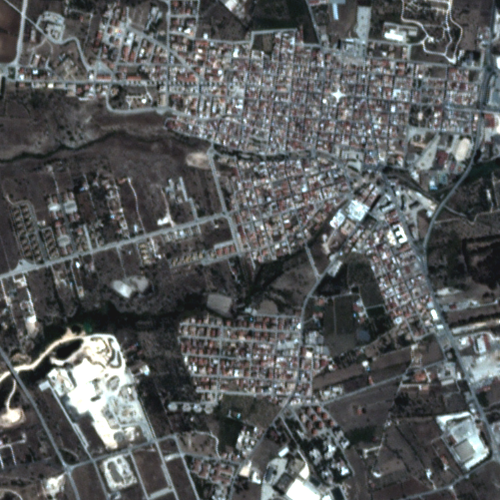}}
\end{subfigure}
\def\pos{(-0.08,1.03)}
\begin{subfigure}{0.46\linewidth}
	\stackinset{r}{}{b}{}{\includegraphics[width=1.2cm]{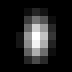}}{\spied{\pos}{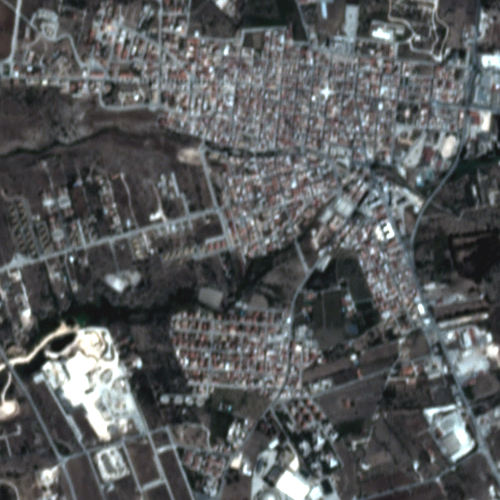}}
\end{subfigure}
\vspace{-0.2em}
\caption{Example of significant blur difference from two consecutive \emph{basic} (non-orthorectified) images from two different satellites. Sharpness scores are $0.036$ and $0.026$. The vertical blur is likely due to an hazardous stabilization of the satellite during the acquisition.}\label{fig:two-basic}
\end{figure}

\vspace{0.5em}\noindent\textbf{Presence of motion blur.~}
Sharpness metrics using MTF is usually calibrated using on-ground targets~\cite{Blanc2009}.
While this allows for very precise estimations, it cannot consider all sources of blur.
In particular, our methodology takes into account resampling as well as blur due to motion during the integration time.
Figure~\ref{fig:two-basic} shows two non-orthorectified images taken on two consecutive days from different satellites. The right image shows an example of motion blur.
The sharpness score allows to automatically filter out such poor quality images using a simple threshold.

\begin{figure}
	\centering
	\includegraphics[width=.9\linewidth]{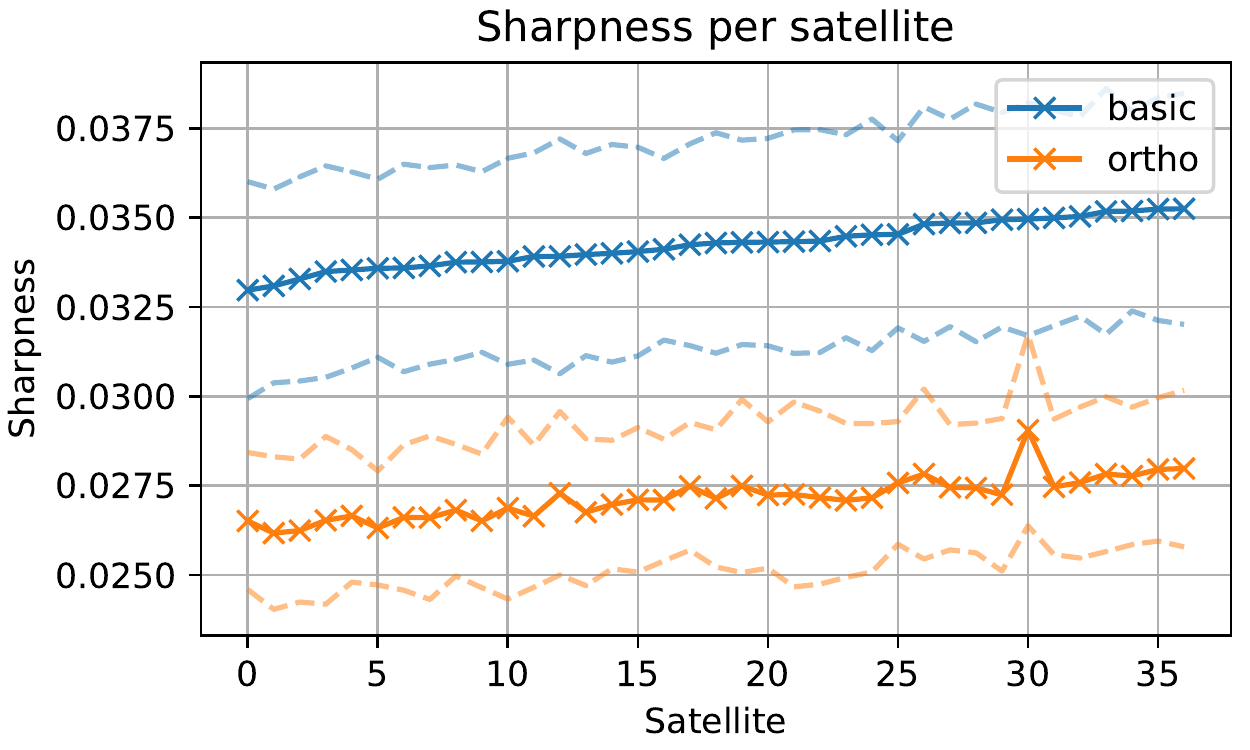}
    \vspace{-.5em}
	\caption{Sharpness mean and variance across the constellation. Satellite ids (in abscissa) were sorted by mean sharpness. Dash-lines indicates the standard deviation for each satellite. The two series correspond to \basic{} and \ortho{} images.}\label{fig:inter-plot-sharpness}
\end{figure}

\vspace{0.5em}\noindent\textbf{Per-satellite sharpness.~}
As previously explained, the PlanetScope constellation is composed of a hundred satellites. Here, 
we study the correlation between the sharpness of the images and the satellite that acquired them.
Our dataset of $14390$ images represent 153 distinct satellites.
In order to have large enough sample sizes, we kept only the satellites for which there are at least 50 images.
Then, we computed the $\sharpness$ score of each image and removed from the dataset all invalid images, that is having a sharpness score below $0.028$ and $0.023$ respectively for \basic{} and \ortho{} images.
Sharpness averages and standard deviations for each satellite are reported in Figure~\ref{fig:inter-plot-sharpness}.
We first notice that the average sharpness is not uniform across the constellation, which would indicate that each satellite produces images with slightly different blur than others.
Indeed, an analysis of variance (one-way ANOVA test) rejects the hypothesis of equal averages and indicates a statistically significant difference in the per-satellite sharpness averages.
Moreover, the clear correlation between the sharpness of \emph{basic} and \emph{ortho} images per satellite confirms that our measure is reliable.
Finally, it is important to note that the standard deviation is large and thus the satellite is not the only factor responsible for the variation of sharpness, and other factors such as motion during acquisition also introduce variance to the effective sharpness of the images.

\section{Conclusion}

In this study, we quantified the variability of blur from PlanetScope images using an efficient blur kernel estimation method and automatically assigning a measure of sharpness to each blur kernel.
The method is blind and does not require the specifications of the optical system.
We also demonstrated that it is possible to apply blind deblurring methods to satellite images in order to equalize quality across time or improve visualization.

Our study of the constellation indicated variation across the images. We observed that the images can contain significant motion blur.
Furthermore, we showed that the orthorectification provided by Planet does decrease the average sharpness of the images.
We also showed correlation between a given satellite and its average sharpness.
Finally, we proposed simple thresholds that allow to discard unsatisfactory images.

However, our method has a few limitations we would like to overcome in future works. First, as explained in Section~\ref{sec:sharpness}, the method is affected by clouds. %
One way to solve this issue would be to apply a cloud detector on the images and mask out detected regions during the kernel estimation.
The second limitation is noise which can be present in some images due to atmospheric conditions and degrades the performance of both kernel estimation and non-blind deconvolution.
We would also like to tackle this problem and provide an additional measure to indicate the noise level and the amount of details we can expect from the restoration.
Finally, saturated region in the image tends to mislead the kernel estimation towards a delta, and further work is required to handle this degradation.

{\small
\bibliographystyle{IEEEbib}
\bibliography{pl}
}

\end{document}